\documentclass{mva_style}
\usepackage{graphicx}
\usepackage{subfig}
\usepackage{multirow}
\usepackage[pagebackref=true,breaklinks=true,letterpaper=true,colorlinks,bookmarks=false]{hyperref}
\usepackage{multicol}
\usepackage{hyperref} 
\usepackage[hyphenbreaks]{breakurl}
\usepackage{tabularx}
\newcommand\Tstrut{\rule{0pt}{2.6ex}}

\finalcopy %Uncomment this line for the Camera-Ready Manuscript

\begin{document}

% {\title{FoodTracker: A Real-time Food Detection Mobile Application by Deep Convolutional Neural Networks}\thanks{Just a note.}}

\title{FoodTracker: A Real-time Food Detection Mobile Application by Deep Convolutional Neural Networks\thanks{Accepted by The 16th International Conference on Machine Visiaon Applications, Tokyo, Japan, 2019}}

\author{Jianing Sun, Katarzyna Radecka, Zeljko Zilic\\
Department of Electrical and Computer Engineering, McGill University\\
Montreal, Quebec, Canada\\
{\tt\small jianing.sun@mail.mcgill.ca, {\{katarzyna.radecka, zeljko.zilic\}@mcgill.ca}}
}

\maketitle

\section*{\centering Abstract}
\textit{
  We present a mobile application made to recognize food items of multi-object meal from a single image in real-time, and then return the nutrition facts with components and approximate amounts. Our work is organized in two parts. First, we build a deep convolutional neural network merging with YOLO, a state-of-the-art detection strategy, to achieve simultaneous multi-object recognition and localization with nearly 80\% mean average precision. Second, we adapt our model into a mobile application with extending function for nutrition analysis. After inferring and decoding the model output in the app side, we present detection results that include bounding box position and class label in either real-time or local mode. Our model is well-suited for mobile devices with negligible inference time and small memory requirements with deep learning algorithm. 
}

\section{Introduction}
\label{1}
With people paying increasing attention to tracking meal items and nutrition contents in daily life for weight loss or medical purposes, self-managed food monitoring has become one of the most practical application of computer vision. 
However, most current mobile apps (CarbandMove, MyFitnessPal, FatSecret, etc) require manual data entry, which is tedious and time consuming. Cordeiro~\textit{et. al}~\cite{apps} conducted a survey in food journaling in 2015. They find that even if 80\% (117/141) food journalers reported archiving data with mobile apps, most users do not use such apps for very long. 

To avoid manual data recording, some recent work~\cite{zhang} use computer vision techniques to reason about meals, but they only work in laboratory conditions with well-separated food items and the number of categories is small. Google proposed a system called Im2Calories~\cite{Im2Calories} with a mobile app that can recognize a food object in an image, and then predict its nutritional content based on volume estimation result. However, they rely on the menu database and restaurant database collected from a specific area, and can only recognize a single food object in an image. They generate segmentation mask instead bounding box, and this part is presented only in laboratory environment with few categories. 

The ideal system in food journaling is a real-time automatic system including localization, recognition, and precise volume estimation generated from solely image input. However, even though there is a large number of prior work~\cite{tada}~\cite{tada2} in this area, food journaling is still widely considered difficult. The challenges stem from causes such as: the vast range of intraclass variations, the components complexity in many foods, e.g. sandwich, and the huge amount of categories and regional specialty.

In this paper, we propose an approach, which utilizes several state-of-the-art deep learning techniques, trained to detect\footnote{Following common terminology, we denote by~\textbf{detect} a combination of localize and recognize.} multiple food items in real-time. Our system is tailored to run on a conventional mobile phone with built-in camera, and presented with nutrition facts for each food item served in an unit amount. 

We rely on our prior work~\cite{sun} producing a deep convolutional neural network (DCNN) model built upon Mobilenet~\cite{mobilenet}, adapting with a state-of-the-art~\emph{one-stage} detection framework, YOLOv2~\cite{yolov2} to generate bounding box and class label simultaneously. The network structure is small and computationally inexpensive thanks to the leverage of depth-wise separable convolution~\cite{depthwise} and YOLOv2 strategy. Next we fit our deep learning model into a mobile application by using the TensorFlow Java API~\cite{tf}, and decode our model output with fast inference with real-time user image input. Finally, we present detection results with nutrition contents in the app. 

This paper is structured as follows. Section~\ref{2} explains in detail how depth-wise separable convolution achieves less computation cost and parameter reduction. Section~\ref{3} describes how we process the training, the DCNN architecture and the YOLO strategy. Section~\ref{4} illustrates the mobile application prototype with the user interface outline. Section~\ref{5} closes with a discussion and a summary of the future work.

%------------2 Preliminary-------------------------------
\section{Background}
\label{2}
\label{2.1}
\subsection{Depthwise Separable Convolution}
Our DCNN structure is built upon MobileNet, which is designed for mobile and embedded vision applications. MobileNet is based on a streamlined architecture that utilizes depth-wise separable convolutions~\cite{depthwise} to construct lightweight deep neural networks. A depth-wise separable convolution block factorizes a standard convolution into a depth-wise convolution and a 1x1 point-wise convolution. Compared to a standard convolution layer that filters and combines input feature maps into a new set of outputs in one step, depth-wise separable convolution splits this process into two layers, with a separate layers for filtering and combining.

By expressing standard convolution as a two-step process, we achieve substantial reduction in both the parameter count and the computation cost. 

More precisely, in a standard convolution layer, we denote by $D_F$ the spatial dimension of the input feature map, $D_K$ the filter width and height (assumed to be square), $M$ the number of input channels and $N$ the number of output channels. We also assume~\texttt{SAME} padding so that the output feature map has the same spatial dimensions as the input.

Then, the parameter number ratio between depth-wise separable convolution and standard convolution is:
$$\frac{D_K \cdot D_K \cdot M + M \cdot N}{M \cdot N \cdot\ D_K \cdot D_K}  =  \frac{1}{N} + \frac{1}{D_K^2}$$

Likewise, the computation cost ratio between depth-wise separable convolution and standard convolution:
$$\frac{D_K \cdot D_K \cdot M \cdot D_F \cdot D_F + M \cdot N \cdot D_F \cdot D_F}{D_K \cdot D_K \cdot M \cdot N \cdot D_F \cdot D_F} = \frac{1}{N} + \frac{1}{D_K^2}$$

MobileNet uses depth-wise separable convolutions with 3x3 filter size, leading to 8 to 9 times fewer computations than the standard convolutions, with only a small reduction in accuracy~\cite{mobilenet}.

%------------3 Food Detection------------------
\section{Food Detection}
\label{3}
In this section, we first describe how to prepare the training instances, outline the principle of YOLO, then illustrate the overall DCNN architecture and present the mean average precision (mAP) result evaluated on UECFood100~\cite{uecfood100} and UECFood256~\cite{uecfood256} benchmarks. These two datasets contain around 40K images (mostly Japanese food items) in total with 100 and 256 categories, respectively.

%-----------3.1 Data Augmentation ---------------

\begin{figure}[t]
    \begin{center}
    \vspace{0.5em}
    \includegraphics[width=0.95\linewidth, height=14cm]
    {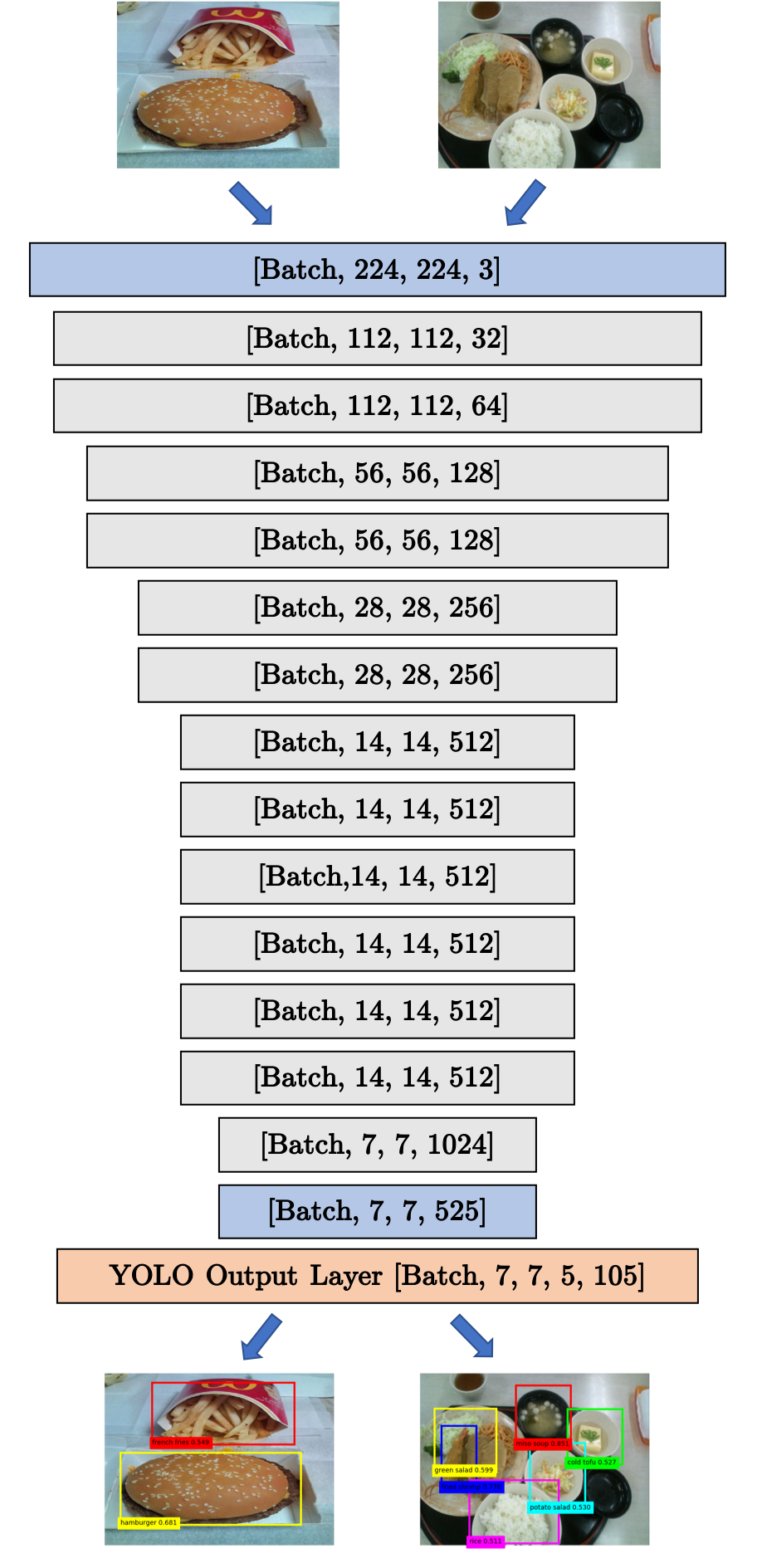}
    \vspace{-1em}
    \end{center}
    \caption{MobileNet-YOLO Architecture. Each grey triangle represents a convolution block. Blue triangle represents standard convolution layer. \emph{Batch} is the number of training instances in each iteration.}
    \vspace{-1.5em}
\label{fig:dcnn}
\end{figure}

\subsection{Data Augmentation}
\label{3.3}
Data augmentation represents the process of artificially increasing the number of training instances~\cite{da}. Previous research demonstrates that data augmentation can act as a regularizer in preventing overfitting in neural networks~\cite{da2}. Our DCNN model aims to be applied in mobile devices with camera input in real-time, which can vary in illumination, viewpoints, backgrounds, etc. Thus we utilize several data augmentation methods to improve the robustness and alleviate overfitting. 

In particular, half of the training instances are processed with one of the following data augmentation methods: blur, horizontal flip, Gaussian noise or color shift.
Although the training time is increased by 5 to 6 times due to data augmentation, our model is more robust and less overfitting, which is critical and beneficial for real-time image processing.

%------------3.2 YOLOv2 --------------------------
\subsection{YOLOv2 Strategy}
\label{3.2}
After downsizing the input resolution from (224, 224) to (7, 7) by 13 depth-wise separable convolution blocks in MobileNet, we add one more convolution layer reshaping the output size to be compatible with the YOLO output layer. 

The main idea here is to divide each input image into (S, S) grid, and predict N bounding boxes in each cell with the shape $[b_x, b_y, t_w, t_h]$ for a bounding box along with a box confidence score for each box. In this case, $t_w, t_h$ are offsets for the N prior anchor boxes generated by clustering.

\begin{figure}[h]
    \centering
    \subfloat[]{{\includegraphics[width=4cm,height=3.7cm]{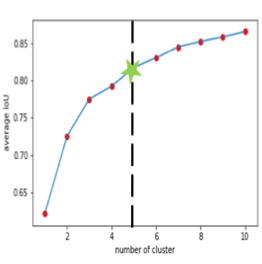} }}
    \hspace{1mm}
    \subfloat[]{{\includegraphics[width=3.5cm,height=3.5cm]{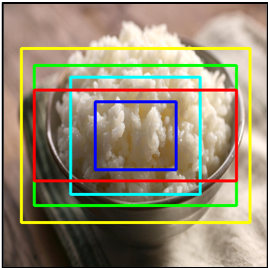} }}%
    \caption{(a)K-means clustering box dimensions on UECFood100. (b)Prior anchors illustration in an example image.}
    \vspace{-1em}
    \label{fig:knn}%
\end{figure}

To find out the suitable number of clusters and the prior shapes of bounding boxes predicted in each cell, we use k-means cluster algorithm~\cite{knn_algorithm} to generate prior anchors' shapes with respect to the cluster number under the Intersection over Union (IoU) metric. IoU is simply computed as the ratio of the area where the two boxes overlap over the total union area of the two boxes. We compute IoU for each KNN-generated anchor with the ground truth box, then we compute the average IoU under different cluster numbers from 1 to 10 to balance between number of anchors with model complexity as shown in Fig.~\ref{fig:knn}(a). 

Even if more anchors lead to higher average IoU, the model complexity will also increase, giving rise to higher computation cost with little benefit in performance. We find that k=5 gives a good trade-off for average IoU vs. model complexity, and we elaborate the five k-means generated prior anchors in Fig.~\ref{fig:knn}(b).

%------------3.2 DCNN Structure-------------------
\subsection{DCNN Architecture}
\label{3.3}
All convolution layers are followed by a batchnorm~\cite{bn} and ReLU nonlinearity~\cite{relu} with an exception of the final fully connected (FC) layer that has no nonlinearity, and feeds into a YOLO output layer. Counting depth-wise and point-wise convolutions as separate layers, our MobileNet-YOLO architecture has 30 layers with 3.5 million parameters in total. The overall scheme is shown in Fig.~\ref{fig:dcnn} with two example inputs with results. The input resolution for training is (224, 224).

For training, we use Adam~\cite{adam} optimizer with $lr=1e-4$, reduce factor = 0.5, $\beta_1=0.9, \beta_2=0.999, \epsilon=1e-8$, batch size = 16, with 11K iterations in total. 

We evaluate our model on UECFood100~\cite{uecfood100} and UECFood256~\cite{uecfood256} sets with validation on around 10K food images. We achieve satisfying food detection performance with \textbf{76.36\%} mAP on UECFood100 and \textbf{75.05\%} on UECFood256 with \texttt{IoU=0.5} on validation set, hence our model is well-suited to be fit into mobile devices. 

\begin{figure}[t]
    \begin{center}
    \includegraphics[height=0.8\linewidth]
    {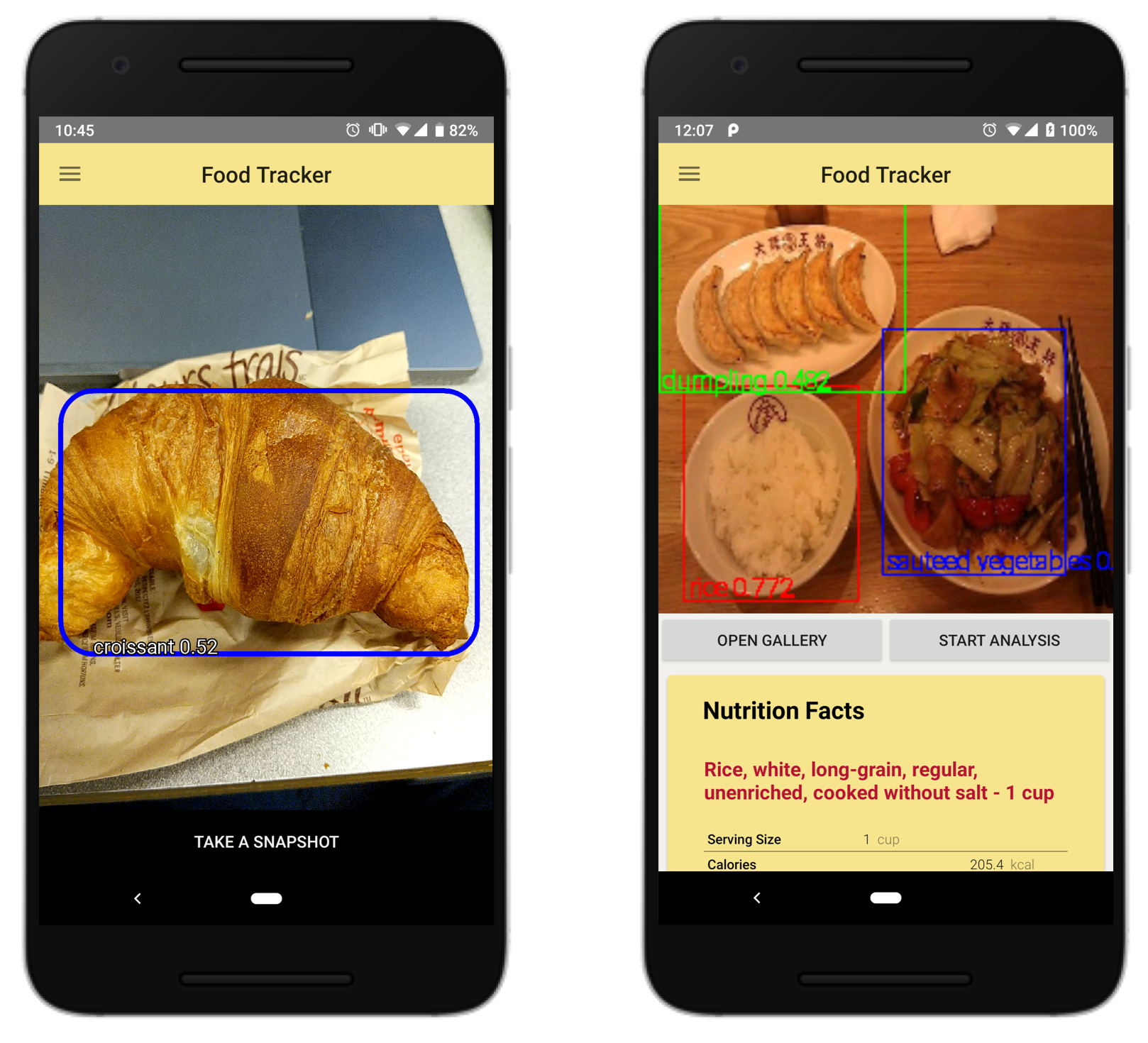}
    \end{center}
    \caption{\textbf{Left:} real-time food detection example with results. \textbf{Right:} local mode food detection and nutrition analysis, user can select an picture for processing from local gallery. }
\label{fig:2app}
\end{figure}

%----------4 Mobile Application -------------------
\section{Mobile Application}
\label{4}
We convert our trained model from \texttt{.h5} file to \texttt{.pb} file so that our model can make prediction by using TensorFlow Java library~\cite{tf}. 

For each inference, we decode the output and select the bounding box with confidence score higher than 0.4, locate boxes by $b_x, b_y$ and adjust them based on the prior anchors generated in Section~\ref{3.2} with the predicted offsets $t_w, t_h$. Moreover, we perform non-maximal suppression~\cite{yolo} to get rid of duplicate boxes by checking if there are overlapping with predicted boxes with more than 30\% area. 

% suppress non-maximal boxes~\cite{yolo} by checking if they are duplicates with other boxes with YOLO's parameter \texttt{nms\_threshold=0.3}. 

\begin{figure}[t]
    \begin{center}
    \vspace{0.5em}
    \includegraphics[width=0.48\textwidth]
    {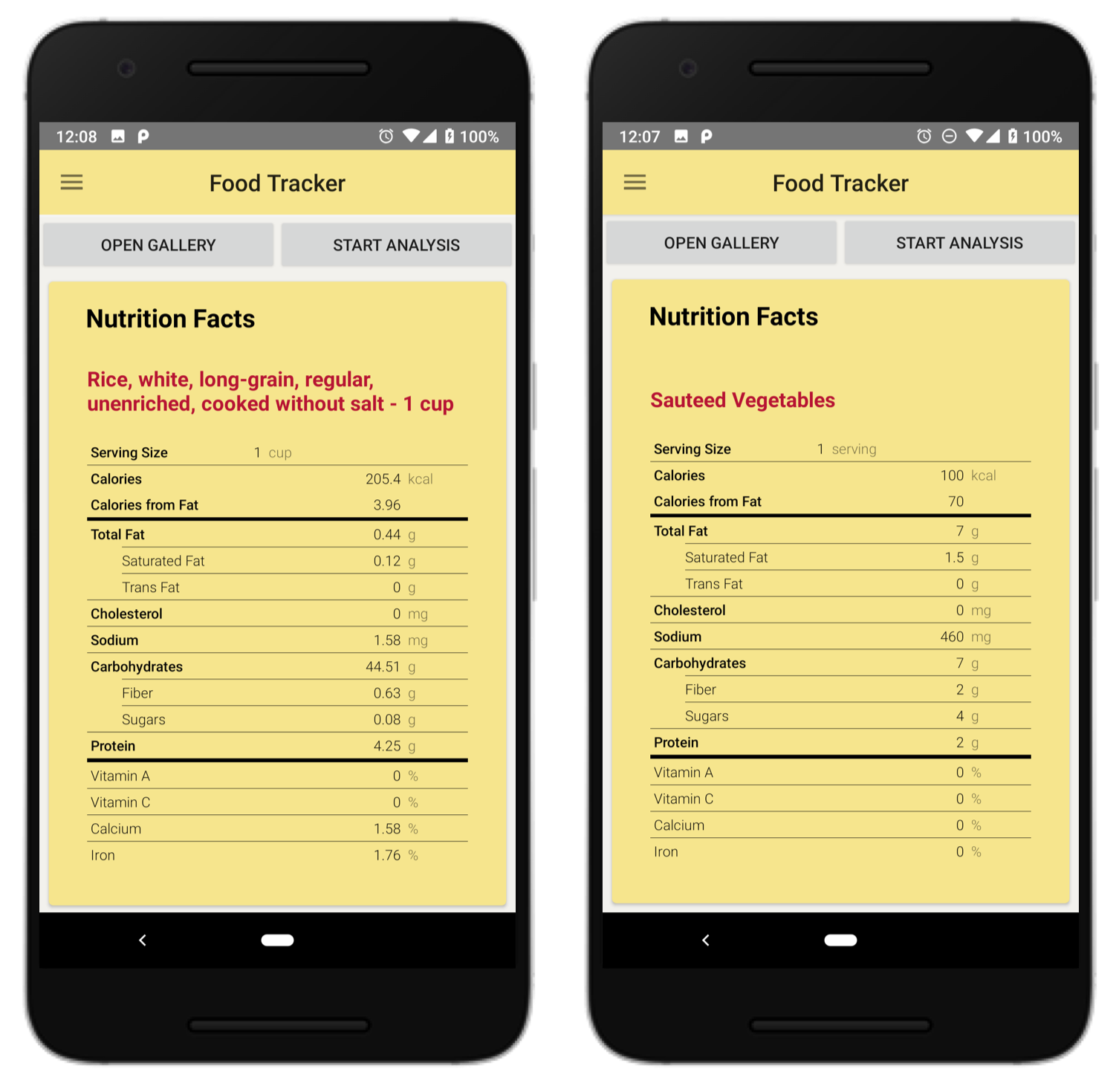}
    \end{center}
    \caption{Example of a multi-object nutrition analysis user interface based on the food detection results. }
    \vspace{-0.5cm}
\label{fig:3app}
\end{figure}

Considering that our datasets contain mostly Asian food, which is usually served in a container (similar to western fast food), we assume, that the amount of each detected food item is one serving. Our nutrition analysis function is inspired by NutriVision~\cite{nutrivision}, which utilizes one of the biggest global nutrition databases, Nutritionix~\cite{nutri}, to analyze the nutrition content. Nutritionix contains more than 700K food items ranging from common foods, restaurant dishes and grocery foods. The deficiency of NutriVision is it requires manual data entry. We call the Nutritionix API based on our trained DCNN model's output, and return the one serving food item nutrition facts on the user interface, as shown in Fig.~\ref{fig:3app}.

%, though it requires manual data entry.

There are two modes in our mobile app design, as shown in the left and right side of Fig.~\ref{fig:2app}. One is the real-time mode to fast infer the camera input per frame and display the detection result including bounding box, class label and confidence score in the screen. Second is the local mode where users can select an image from local gallery to do nutrition analysis along with food detection. 

We do not rely on any remote server to do the computation required for image inference. The inference process is conducted solely in the mobile device. We test our mobile app on a Google Pixel 2 with Android 9.0 system and a OnePlus 5 with Android 8.0 system in both real-time and local mode. Thanks to our small neural network design and efficient detection strategy, our average wall clock time\footnote{Wall clock time is the actual time taken by a computer to complete a task. It is the sum of CPU time, I/O time, and the communication channel delay.}  per image is \textbf{75ms} and average CPU time\footnote{CPU time measures only the time during which the processor is actively working on a certain task.} per image is \textbf{15ms} in real time (Table.~\ref{table:profile}), thereby there is almost no delay observed by users. 
%Output snapshots are shown in Fig.~\ref{fig:2app}.

%--------Discussion and Future Work--------------
\section{Discussion and Future Work}
\label{5}
The process of automatically detecting food objects and extracting nutrition contents is very complicated, especially when it comes to real life application. In particular, it requires solving various problems, such as: fine-grained recognition to distinguish subtly different forms of food, instance segmentation and counting, mask generation, depth/volume estimation from a single image. Most of the existing state-of-the-art work focuses specifically on one of the sub-problems of food detection with computer vision techniques~\cite{discussion1}~\cite{discussion2}~\cite{discussion3}. They all focus on a single task with strict environmental conditions or external assistance so that they are still far away from the holy grail of the automated food journaling systems. 

 Mobile application development is one of the most promising areas because of wide popularity and usability of smartphones. On the other hand, the limitation of computational resources of mobile devices makes it difficult to apply deep learning techniques. As for food detection and nutrition analysis with computer vision, many research conducted in laboratory environment making use of the power of multiple graphic cards, fewer of them have been generalized to mobile devices, still among those even fewer have been made publicly available. 

For all these aforementioned problems, our work has tackled some of these, but it is clear that there is much more work to do. We believe that multi-task learning in this area is essential as the process from image to nutrition is involved with many computer vision problems. Moreover, the supervise learning attribute of most computer vision algorithms give rise to some limitations on the range of testing categories. Also, the necessary of ground truth annotation has become a trouble, especially for food-related task with countless food categories around the world. 

In research to follow, we will be working on mask generation with respect to volume estimation, and investigate how to better leverage deep learning into mobile devices taking the computational resource limitation into consideration. There are many details to handle and many interesting problems from the point of view of computer vision research. Nevertheless, we believe that even a partial solution to these problems could be of great value to the society. 
% Further, we will release the mobile app shortly in Google Play.

%------------Reference-------------------------------
\begin{table}[t]
\caption{App performance profile.}
\vspace{-1.5em}
    \begin{center}
        \begin{tabular}{cccc}
            \hline\hline 
            % & \\[\dimexpr-\normalbaselineskip+5pt]
            \multicolumn{2}{l|}{\multirow{2}{*}{Inference time}} & CPU time & 15ms \Tstrut \\ [1ex]
            \multicolumn{2}{l|}{} &Wall clock time & 75ms \\ [1ex]
            \hline
            & \\[\dimexpr-\normalbaselineskip+5pt]
            \multicolumn{3}{c}{DCNN model size} & 8.1MB\\ [1ex]
            \multicolumn{3}{c}{Runtime memory} & 242.2MB\\ [0.5ex]
            \hline\hline
        \end{tabular}
        \vspace{-1.5em}
    \end{center}
\label{table:profile} 
\end{table}

{\small
\bibliography{reference_2}
\bibliographystyle{unsrt}
}

\end{document}